\documentclass{article}
\usepackage{spconf,amsmath,graphicx}

\usepackage{amssymb}
\usepackage{verbatim}
\usepackage{multirow}
\usepackage{comment}

\title{MOTION REPRESENTATION USING RESIDUAL FRAMES WITH 3D CNN}
\name{Li Tao\qquad Xueting Wang\qquad Toshihiko Yamasaki}
\address{The University of Tokyo\\
\{taoli, xt\_wang, yamasaki\}@hal.t.u-tokyo.ac.jp}
\begin{document}
%
\maketitle
\begin{abstract}
  Recently, 3D convolutional networks (3D ConvNets) yield good performance in action recognition. However, optical flow stream is still needed to ensure better performance, the cost of which is very high. In this paper, we propose a fast but effective way to extract motion features from videos utilizing residual frames as the input data in 3D ConvNets. By replacing traditional stacked RGB frames with residual ones, 35.6\% and 26.6\% points improvements over top-1 accuracy can be obtained on the UCF101 and HMDB51 datasets when ResNet-18 models are trained from scratch. And we achieved the state-of-the-art results in this training mode. Analysis shows that better motion features can be extracted using residual frames compared to RGB counterpart. By combining with a simple appearance path, our proposal can be even better than some methods using optical flow streams.
\end{abstract}
\begin{keywords}
Motion representation, action recognition, residual frames, 3D CNN~\footnote{© 2020 IEEE. Personal use of this material is permitted. Permission from IEEE must be obtained for all other uses, in any current or future media, including reprinting/republishing this material for advertising or promotional purposes, creating new collective works, for resale or redistribution to servers or lists, or reuse of any copyrighted component of this work in other works.}
\end{keywords}
\section{Introduction}
\label{sec:intro}

For action recognition, motion representation is an important challenge to extract motion features among multiple frames. Various methods have been designed to capture the movement.
2D ConvNet based methods use interactions in the temporal axis to include temporal information~\cite{karpathy2014large,wang2016temporal,li2019temporal,tsm,wang2018non}. 3D ConvNet based methods improved the recognition performance by extending 2D convolution kernel to 3D, and computations among temporal axis in each convolutional layers are believed to handle the movements~\cite{c3d,p3d,s3d,i3d,res3d,r3d}.
State-of-the-art methods showed further improvements by increasing the number of used frames and the size of the input data as well as deeper backbone networks~\cite{slowfast,tran2019video}.

\begin{figure}[t]
  \begin{center}
  \includegraphics[width=\linewidth]{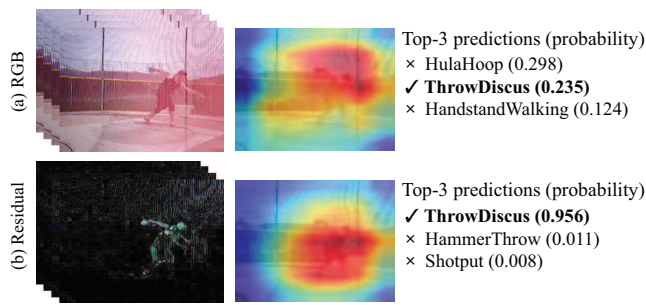}
  \vspace{-20pt}
  \end{center}
     \caption{An example of our residual frames compared with normal RGB inputs. The residual-input model focused on the movement part while RGB-input model paid more attention on background, which lead to lower accuracy for prediction.}
  \label{fig:res_frame}
  \vspace{-10pt}
\end{figure}

In a typical implementation of 3D ConvNets, these methods used stacked RGB frames (or called video clips, we use both in the following descriptions) as the input data. However, this kind of input is considered not enough for motion representation because the features captured from the stacked RGB frames may pay more attention to the appearance feature including backgrounds and objects rather than the movement itself, as shown in the top example in Fig.~\ref{fig:res_frame}. Thus, combining with an optical flow stream is necessary to further represent the movement and improve the performance, such as the two-stream models~\cite{feichtenhofer2016convolutional,feichtenhofer2016spatiotemporal,simonyan2014two}. However, the processing of optical flow greatly increases computation. Besides, two-stream results activation of the optical flow stream can only be obtained after the optical flow data are extracted, which causes high latency.

In this paper, we propose an effective strategy based on 3D convolutional networks to pre-process RGB frames for the generation and replacement of input data. Our method retains what we call \textbf{residual frames}, which contain more motion-specific features by removing still objects and background information and leaving mainly the changes between frames. Through this, the movement can be extracted more clearly and recognition performance can be improved, as shown in the bottom sample in Fig.~\ref{fig:res_frame}. Our experiments reveal that our approach can yield significant improvements over \text{top-1} accuracies when those ConvNets are trained from scratch on UCF101~\cite{ucf101} and HMDB51~\cite{hmdb} datasets. 

For some specific category pairs such as \textit{Playing Guitar} and \textit{Playing Ukelele}, the movements are highly similar while the instruments are different, guitar or ukelele. In this case, it is difficult to distinguish by only motion representation without enough appearance features. Therefore, we propose a two-path solution, which combines the residual input path with a simple 2D ConvNet to extract appearance features from a single frame. Experiments show that our proposed two-path method obtains better performance over some two-stream models on UCF101 / HMDB51 datasets when using the \textbf{same input sizes} and similar or even \textbf{shallower} network architectures. 

Our contributions are summarized as follows:
\begin{itemize}
  \setlength{\parskip}{-3pt}
\item We are the first to use residual frames with 3D ConvNets for action recognition, which is simple, fast, but effective. 
\item Analysis indicates that our proposal can extract better motion representation for actions than RGB counterparts. 
\item Our proposal can achieve the state of the art when models are trained from scratch on two benchmarks. Our results can even achieve better performance with less computation cost than some methods using optical flow.
\end{itemize}


\section{Proposed method}
\label{sec:format}

\subsection{Residual frames}

When subtracting adjacent frames to get a residual frame, only the frame differences are kept. In a single residual frame, movements exist in the spatial axis. Using residual frames with 2D ConvNets has been attempted and proved to be somewhat effective~\cite{wu2018compressed, wang2016temporal}. However, because actions or activities are complex with much longer durations, stacked frames are still necessary. In stacked residual frames, the movement does not only exist in the spatial axis, but also in the temporal axis, which is more suitable for 3D ConvNets because 3D convolution kernels will process data in both spatial and temporal axes. Using stacked residual frames helps 3D convolutional kernel to concentrate on capturing motion features because the network does not need to consider the appearance information of objects or backgrounds in videos. 

Here we introduced the detail calculation of proposed residual frames input for 3D ConvNets. We use $frame_i$ to represent the $i_{th}$ frame data, and $Frame_{i\sim j}$ denotes the stacked frames from the $i_{th}$ frame to the $j_{th}$ frame. The process to get residual frames can be formulated as follows,

\vspace{-15pt}
\begin{equation}
   ResFrame_{i\sim j} = | Frame_{i\sim j} - Frame_{i+1\sim j+1} |
   \vspace{-5pt}
\end{equation}
The computational cost is cheap and can even be ignored when compared with the network itself or optical flow calculation. With this change, 3D ConvNet can extract motion features by focusing more on the movements in videos while ignoring some unnecessary objects and backgrounds. 

We also pay attention to some cases that similar movements could exist in different actions, where only good motion representation is not enough. For example, for actions \textit{Apply Eye Makeup} and \textit{Apply Lipstick}, the main difference lies in the location (around the eye or mouth) of the similar movement. In this example, 3D ConvNets may be able to distinguish them to some extent but the loss of appearance information does increase the difficulty. Therefore, we use a 2D ConvNet to process the lost appearance information and combine with a 3D ConvNet using residual frames as input to form a two-path network.

\subsection{Two-path network}

\begin{figure}[t]
    \begin{center}
    \includegraphics[width=\linewidth]{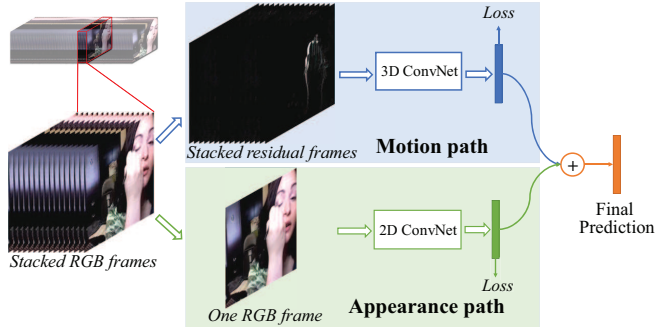}
    \end{center}
    \caption{Framework of our two-path network. The motion path and the appearance path are trained separately using cross-entropy loss. In inference period, the output probabilities from two paths are averaged.}
    \label{fig:twopath}
 \end{figure}

To distinguish our proposal from the existed two-stream methods~\cite{feichtenhofer2016convolutional,feichtenhofer2016spatiotemporal,simonyan2014two}, we refer to our method as `two-path' because we do not use any pre-computed motion features such as optical flow. Our two-path network is formed by a motion path and an appearance path, which is illustrated in Fig.~\ref{fig:twopath}

\noindent{\bf Motion path}.
Because residual frames are used in this path, movements then exist in both spatial axis and the temporal axis. The convolutional kernel for each 3D convolutional layer is in 3 dimensions. For each 3D convolutional layer, data will be computed among three dimensions simultaneously. Therefore, 3D convolutional layers are used in this path. Because there are many existing 3D convolution based network architectures which have been proved effective in many action recognition datasets, we do not focus on designing a new network architecture in this paper. To verify the robustness and versatility of our proposal, we conduct experiments on various models, such as ResNet-18-3D~\cite{res3d}, R(2+1)D~\cite{r3d}, I3D~\cite{i3d}, and S3D~\cite{s3d}.

\noindent{\bf Appearance path}.
By using residual frames with 3D ConvNets, motion features can be better extracted, while background features which contains object appearances are lost. Here, we simply use a naive 2D ConvNet which treats action recognition as a simple image classification problem. This path is a supplementary of motion path.

For the combination of these two paths, we average the predictions for the same video sample. 

\section{Experiments}
\label{sec:pagestyle}

\subsection{Datasets}
 We mainly focus on the following benchmarks: UCF101~\cite{ucf101}, HMDB51~\cite{hmdb}, and Kinetics400~\cite{kinetics}. UCF101 consists of 13,320 videos in 101 action categories. HMDB51 is comprised of 7,000 videos with a total of 51 action classes. Kinetics400 consists 400 action classes and contains around 240k videos for training, 20k videos for validation and 40k videos for testing. We mainly conduct our experiments on UCF101 and HMDB51. Results on Kinetics400 will also be reported to prove the effectiveness of our proposal.

\subsection{Implementation details}
\noindent{\bf Motion path.} In this path, stacked residual frames are set as the network input data. Residual frames are used identically to traditional RGB frame clips. For 3D ConvNets in action recognition, when ignoring the image channel number, $T \times H \times W$ is used to denote the data shape, where $T$ frames are stacked together with height $H$ and width $W$. There are several setting choices for input data shape, such as $16\times112\times112$, $64\times112\times112$, $16\times224\times224$, and $64\times224\times224$. For fair comparison, in all of our motion path, following~\cite{c3d}, frames are resized to $170\times 128$ and $16$ consecutive frames are stacked to form one clip. Then, random spatial cropping is conducted to generate an input data of size $16\times 112\times 112$. Random horizontal flipping and Jittering are also applied during training. We tried two variants of ResNet-18-3D. In~\cite{res3d}, models are directly from image classification tasks. However, the height and weight for video clips are both 112, which is half of 224. Therefore, we delete the pooling layer after the first convolution layer. R(2+1)D~\cite{r3d}, I3D~\cite{i3d}, and S3D~\cite{i3d} are also reimplemented to verify the robustness of our proposal. The batch size is set to 32. When models are trained from scratch, the initial learning rate is set to 0.1. We trained models for 100 epochs on UCF101 and HMDB51. When fine-tuning on UCF101 and HMDB51 using Kinetics400 pre-trained models, model weights are directly from~\cite{res3d} and the network architecture remains the same as~\cite{res3d}. The initial learning rate became 0.001, and 50 epochs were sufficient.

\noindent{\bf Appearance path.} Our appearance path is just a supplemental to our motion path. Therefore, we make it simple and treat action recognition as image classification. The goal for this path is to capture appearance features for background and objects. This progress is standard in image classification to enable the use of ImageNet pre-trained models. ResNeXt-101~\cite{resnext} is used in this path.

\noindent{\bf Testing and Results Accumulation.} 
For the motion path, 16 clips are uniformly sampled from one video regardless of the video length. The predictions are averaged over all video clips to generate the final result. For the appearance path, 16 frames are sampled to match the motion path. And predictions are averaged to generate the final results.

\section{Results and discussion}
\label{sec:typestyle}

In this session, results on motion path are mainly reported because appearance path is only a supplemental part to the motion path. We try our best to make fair comparisons with previous methods. Therefore, for all the comparative methods using 3D convolution, inputs in size $16\times112\times112$ will be reported if available. We do not compare with some methods with state-of-the-art performance such as I3D which used $64\times224\times224$ as input together with optical flow stream due to the large difference on the setting and too high computing cost. Using larger input data and deeper network usually can ensure better performance, while we focus more on the motion representation in this paper. 

\subsection{Effectiveness of residual inputs}
\begin{table}[t]
  \caption{The effectiveness of our residual inputs for scratch training on UCF101 \textit{split}~1. We reimplement R(2+1)D, I3D and S3D, and keep the input in the same shape. In the column \textbf{residual}, if checked, the models are using proposed residual clips, otherwise using original RGB clips as input.}
  \begin{center}
  \scalebox{0.85}{
 \setlength{\tabcolsep}{1.8mm}{
 \begin{tabular}{c c c c}
    \hline
    Model & \textbf{residual} & top-1 & top-5 \\
    \hline
    ResNet-18 (baseline) & $\times$ & 51.9 & 76.3 \\
    ResNet-18 (baseline) & \checkmark  & 66.4 & 88.0 \\
    ResNet-18 (delete first pooling layer) & $\times$  & 61.6 & 84.9 \\      
    ResNet-18 (delete first pooling layer) & \checkmark & \textbf{78.0} & \textbf{94.0} \\
    \hline
    R(2+1)D~\cite{r3d} & $\times$ & 51.8 & 79.2 \\
    R(2+1)D\cite{r3d} & \checkmark & \textbf{66.7} & \textbf{88.3} \\
    \hline
    I3D~\cite{i3d} & $\times$ & 56.5 & 81.3 \\
    I3D~\cite{i3d} & \checkmark & \textbf{66.6} & \textbf{87.0} \\
    \hline
    S3D~\cite{s3d} & $\times$ & 51.1 & 77.4 \\
    S3D~\cite{s3d} & \checkmark & \textbf{64.8} & \textbf{86.9} \\
    \hline
  \end{tabular}}}
  \end{center}
\label{table:result_ucf3d}
\end{table}

Significant improvements can be obtained using our residual inputs when trained from scratch. As shown in Table.~\ref{table:result_ucf3d}. In addition to ResNet-18, we also reimplement R(2+1)D, I3D and S3D. By replacing the RGB clips with residual clips, for all models, more than 10\% gain can be achieved. By using our ResNet variant together with residual clips as inputs, the top-1 accuracy can be improved from 61.6\% to 78.0\%. 

\begin{figure}[t]
  \begin{center}
  \includegraphics[width=\linewidth]{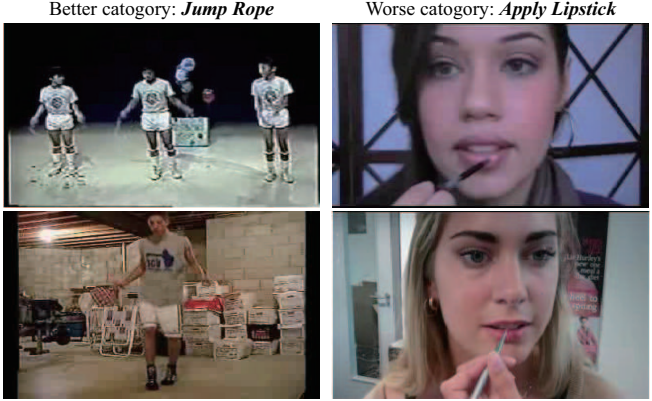}
  \end{center}
     \caption{Examples for case study comparing to RGB-input models. Residual-input model has better performances for category \textit{Jump Rope} while worse for \textit{Apply Lipstick} than RGB-input model because it is more robust with \textbf{mess background} while easy to be confused with \textbf{similar movements} (\textit{Apply Eye Makeup}).}
  \label{fig:cases}
\end{figure}

We also conduct some case study to see what kind of feature the model has learned. In Fig.~\ref{fig:cases}, we can find that for category \textit{Jump Rope}, movements in different samples are in consistent while backgrounds vary from one to another. Our residual-input model can handle these cases easily. Just because the residual-input model can represent the motion itself well, only using residual-input model can not distinguish different actions with similar movements, such as category \textit{Apply Lipstick} and category \textit{Apply Eye Makeup} shown on the right of Fig. 3. Moreover, visualizations using Grad-Cam~\cite{gradcam} in Fig.~\ref{fig:res_frame} also indicate that, RGB-input model will still pay more attention to the background while our motion path will focus on the movement part.

\begin{table}[t]
  \caption{Accuracy using single path on UCF101 and HMDB51. Results are averaged over 3 splits except for the scratch training part, which are results on \textit{split}~1 only. $^*$ indicates results evaluated using the \textbf{same input size} ($16\times112\times112$) with us, otherwise, the size for single frame will be $4\times$ larger. $^\dag$ indicates methods using optical flow.}
  \begin{center}
  \scalebox{0.9}{
  \setlength{\tabcolsep}{0.45mm}{
 \begin{tabular}{c c c c}
    \hline
     & UCF101 & HMDB51 & Kinetics400\\
    \hline
    \textbf{Scratch training} & & & \\
    ResNet-18 baseline~\cite{res3d}$^*$ & 42.4 & 17.1 & 54.2 \\
    STC-ResNet-101~\cite{stc}$^*$ & 45.6 & - & 64.1 \\
    NAS~\cite{nas}$^*$ & 58.6 & - & -\\
    Slow fusion & 41.3 & - & - \\
    TSN (RGB only)~\cite{wang2016temporal} & 48.7 & - & - \\
    C3D~\cite{c3d}$^*$ & 51.6 & 24.3 & 55.6 \\
    \textbf{Motion path (ours)} & \textbf{78.0} & \textbf{43.7} & - \\
    \hline
    \textbf{Single path (fine-tuning))} & & \\
    CoViAR (Residuals)~\cite{wu2018compressed} & 79.9 & 44.6 & -\\
    TSN (RGB difference)~\cite{wang2016temporal} & 83.8 & - & -\\
    ResNet-18 baseline~\cite{res3d}$^*$ & 84.4 & 56.4 & - \\
    C3D (+SVM)~\cite{c3d}$^*$ & 82.3 & 51.6 & -\\
    TSN (RGB, ImNet pretrain)~\cite{wang2016temporal} & 85.7 & 51.0 & - \\
    I3D (RGB, ImNet pretrain)~\cite{i3d} & 84.5 & 49.8 & 71.1\\
    \textbf{Motion path (ours)} & \textbf{89.0} & \textbf{58.1} & 60.3\\
    \hline
    \textbf{Multi-path (fine-tuning)} & & \\
    Two-stream~\cite{simonyan2014two}$^\dag$ & 86.9  & 58.0 & 65.6 \\
    Two-stream (+SVM)~\cite{simonyan2014two}$^\dag$ & 88.0  & 59.0 & -\\
    I3D~\cite{i3d}$^\dag$ & \textbf{98.0} & \textbf{80.7} & \textbf{74.2} \\
    CoViAR (3 nets)~\cite{wu2018compressed} & 90.4 & 59.1 & - \\
    \textbf{Two-path (ours)} & \textbf{90.6} & 56.6 & 67.7\\
    \hline
  \end{tabular}}}
  \end{center}
  \label{table:results}

\end{table}

\subsection{Comparisons with other methods}
First, we compared our motion path with other methods in the first part of Table~\ref{table:results}. All methods are trained from scratch. NAS~\cite{nas} used network architecture search technology to search better network architecture for action recognition and achieve high performance for scratch training. Our motion path has 23.8\% improvement over it. Our proposal method achieves the state-of-the-art when trained from \textbf{scratch} on UCF101 and HMDB51.

Then we introduce the results of our motion path with fine-tuning comparing to other methods shown in the second part in Table~\ref{table:results}. We are not proposing new network architecture, therefore, weights from~\cite{r3d} are directly used for fine-tuning our motion path to fit residual inputs. Residual frame / frame differences has been tried or used in 2D convolution networks such as CoViAR~\cite{wu2018compressed} and TSN~\cite{wang2016temporal}. We also make an apple-to-apple comparison with them. Results show that by using residual frames with 2D CNN, on UCF101, 83.8\% and 79.9\% points can be achieved for TSN (RGB difference) and CoViAR (residuals) at top-1 accuracy while our motion path can get 89.0\%, which reveals that 3D CNNs are more capable of processing residual frames. I3D can achieve the state-of-the-art result (98.0\% at top-1 accuracy) on UCF101 because the input size is $16\times$ ours and optical flow is added, together with both ImageNet and Kinetics400 knowledge. With only RGB input and knowledge from ImageNet, the result for I3D (RGB) is 84.5\%, and our motion path is better. On HMDB51, the same trend can be found. The proposed motion path is even better than I3D (RGB) even though our input size is smaller. Considering the time cost of scratch training on Kinetics400, we directly fine-tuned the pretrained RGB weights to fit our residual inputs. It is interesting that this also works and the top-1 accuracy for our motion path is 60.3\%. This result is higher than 54.2\%, which is achieved by using ResNet-18-3D model with RGB input in~\cite{res3d}.

As shown in the third part of Table~\ref{table:results}, by using an additional appearance path, results on UCF101 can be further enhanced to 90.6\%, which is higher than CoViAR which even used 3 networks. On Kinetics400, our results are 67.7\%, 2.1\% higher than two-stream method~\cite{simonyan2014two}. Moreover, for two-stream methods, optical flow features need to be extracted first. It takes about 48 seconds for a 6-second video (165 frames) using TV-L1 optical flow algorithm~\cite{tvl1} and OpenCV on CPU. Though it can be accelerated by parallel computing, it is still time consuming compared to the inference time of
our motion path (less than 0.19 second/video). Although our results are lower than I3D~\cite{i3d}, it is acceptable considering the input size and lower cost without need of calculating optical flow. A descent is observed on HMDB51 can be considered that the current path is too naive which does not utilize any temporal information. However, the  still helps boost the result using single motion path by 1.6\% and 7.4\% on UCF101 and Kinetics400, separately.

\section{Conclusion}
\label{sec:refs}

In this paper, we mainly focused on extracting motion features without using optical flow. We improved use of 3D convolution by using stacked residual frames as the network input. Results of our proposal could be improved significantly when trained from scratch on UCF101 and HMDB51 datasets. Analysis implied that residual frames can be a fast but effective way for a network to capture motion features and they are a good choice for avoiding complex computation for optical flow. 

\section*{Acknowledgement}
This work was partially financially
supported by the Grants-in-Aid for Scientific Research Numbers
JP19K20289 and JP19K22863 from JSPS, Japan.

\small
\bibliographystyle{IEEEbib}
\bibliography{refs}

\end{document}